\pdfoutput=1
\documentclass[11pt]{article}

\usepackage[]{acl}

\usepackage{times}
\usepackage{latexsym}
\usepackage{examples}

\usepackage{floatrow}
\newfloatcommand{capbtabbox}{table}[][\FBwidth]
\usepackage{graphicx}
\usepackage{caption}
 \usepackage{multirow}
 \usepackage{makecell}
\usepackage{subcaption}
\usepackage{booktabs}
\usepackage[normalem]{ulem}
\useunder{\uline}{\ul}{}
\usepackage{amsmath}
\usepackage{listings}
\newsavebox{\LstBox}
\DeclareMathOperator*{\argmax}{argmax}
\usepackage[T1]{fontenc}
\usepackage[utf8]{inputenc}

\usepackage{microtype}

\title{One Size Does Not Fit All: \\ The Case for Personalised Word Complexity Models}

\author{Sian Gooding \\
  University of Cambridge \\
  \texttt{shg36@cam.ac.uk} \\\And
 Manuel Tragut \\
  Google \\
  \texttt{mtragut@google.com} \\}

\begin{document}
\maketitle
\begin{abstract}
Complex Word Identification (CWI) aims to detect words within a text that a reader may find difficult to understand. It has been shown that CWI systems can improve text simplification, readability prediction and vocabulary acquisition modelling. However, the difficulty of a word is a highly idiosyncratic notion that depends on a reader's first language, proficiency and reading experience. In this paper, we show that personal models are best when predicting word complexity for individual readers. We use a novel active learning framework that allows models to be tailored to individuals and release a dataset of complexity annotations and models as a benchmark for further research.\footnote{Dataset is available at: \url{https://github.com/siangooding/personal_CWI}}
\end{abstract}

\section{Introduction}
Illiteracy affects at least $773$ million people globally.\footnote{\url{https://en.unesco.org/themes/literacy}} For these individuals, reading information from unfamiliar sources or on unfamiliar topics can be extremely difficult. Furthermore, there are a reported $1.5$ billion English-language learners worldwide, spanning all proficiency levels. Statistics such as these illustrate that the comprehension of written text across the world varies substantially. Even those who are deemed functionally literate have differing degrees of text comprehension. 
There are many characteristics of a written text that impact how difficult it is to read. However, one of the main aspects contributing to overall text difficulty is lexical complexity, i.e.\ word complexity \cite{dubay:2004:theprinciples}. Lexical content has been shown to be more important than syntactic features when predicting readability \citep{vajjala2012improving} and a study performed by \citet{nation:2006} found that a reader needs to be familiar with at least 95\% of the words contained within a text for them to understand the content. \newpage
\begin{examples}\label{ex:cat}
\item \center{The \textit{cantankerous} cat. \\ $\downarrow$} \\ The \textbf{\textit{grumpy}} cat. 
\end{examples}
\vspace{0.5cm}
Complex word identification (CWI) focuses on identifying words in a text that may be difficult for a reader to understand and therefore may benefit from simplification~\cite{shardlow:2013:ACL}. For instance, in example (\ref{ex:cat}) a CWI system may identify \textit{cantankerous} as a complex word, then a lexical simplification system is able to substitute it with a simpler alternative, e.g.\ \textit{grumpy} \cite{paetzold2016benchmarking, gooding2019recursive}.

Many downstream tasks benefit from the identification of complex words. For example, CWI has been shown to improve readability assessment systems~\cite{maddela2018word} as well as vocabulary acquisition modules in educational applications~\cite{zaidi2020adaptive}. Furthermore, CWI can significantly reduce errors in simplification systems~\cite{shardlow2014out} and result in higher quality simplifications~\cite{lee2018personalizing}.

Previous work on modelling CWI, including two shared tasks, has relied on the collection of large corpora containing words that have been annotated for their complexity \cite{paetzold-specia:2016:SemEval2, yimam2018report}. However, annotating the difficulty of words is a subjective task, and previous data collection has yielded low levels of annotator agreement \cite{specia2012semeval,paetzold-specia:2016:SemEval2}. As a way of mitigating individual differences, such datasets typically present a homogeneous view on word complexity, by merging annotations across readers~\cite{gooding2021word}. Further attempts to improve annotator agreement have included offering bonus incentives for annotators who select words matching other annotations \cite{yimam-EtAl:2017:RANLP}, as well as providing guidelines for annotators to mark words that they assume would be complex for audiences such as children and those with learning difficulties. Such incentives and instructions change the motivation of annotators and undermine the validity of the resulting complexity labels \cite{ethical_cwi}. 

It is evident that lexical complexity is a highly subjective and idiosyncratic notion. In this paper, we argue that contrary to the one-size-fits-all paradigm currently dominating many natural language processing tasks, the modelling of individual differences in word complexity is critical. Many of the downstream tasks that benefit from CWI, such as text simplification, would be more effective if tailored to the individual reader. In order to facilitate personalised CWI, and therefore personalised text simplification and readability systems, audience specific complexity annotations are required \cite{bingel2018lexi}. Whilst it has been shown that the concept of word complexity, and thus the level of agreement, is aligned between individuals sharing a common background \cite{gooding,gooding2021word}, we argue that the best CWI model for each individual is trained with them `in the loop'.

In our paper, we use active learning to produce per-individual models of word complexity. We collect complexity annotations from $1{,}225$ English as a second language (ESL) readers. Using a novel active learning framework, we show that individual models of word complexity can be trained without extensive annotation. Our experiments demonstrate that bespoke models outperform baselines and a state-of-the-art system when predicting word complexity for individuals, thereby emphasising the importance of viewing the task at a personal level. We present the following contributions:
\begin{itemize}
    \item We are the first to train personalised word complexity models and show that training can be achieved in real-time using active learning. 
    \item We show that per-individual models outperform baseline and state-of-the-art approaches when predicting word complexity for individuals, most notably for lower proficiency levels. 
    \item We demonstrate the benefits of bespoke CWI models in downstream tasks such as text simplification and proficiency prediction.
    \item We release a novel dataset containing $55{,}125$ English word complexity annotations from $1{,}225$ ESL participants, as well as all trained models, as a benchmark for further research.
\end{itemize}

\section{Background}
\subsection{Complex Word Identification}
The first dataset collected for the evaluation and benchmarking of CWI was the CW corpus \cite{shardlow:2013:PITR}. The dataset contained $731$ instances, collected using simplifications made by human editors when adapting content for Simple Wikipedia articles. This dataset was extended with further Wikipedia examples, including data collected by \citet{hornEtAl:2014:ACL}, for the first CWI shared task. The resulting dataset contained a total of $9{,}200$ sentences which were annotated by $400$ non-native speakers for complex words. The shared task organised by \citet{paetzold-specia:2016:SemEval11} required the prediction of binary word complexity for non-native speakers. However, as the dataset had low annotator agreement, the merged notion of complexity was difficult to model, and all systems resulted in low F-scores. 
%The efficacy of Simple Wikipedia edit histories as a 'gold standard' in text simplification research has been disputed \cite{amancio2014analysis, xu:2015:TACL}, instead this

Following this, \citet{yimam-EtAl:2017:RANLP} collected a CWI dataset which was used in a second shared task \cite{yimam2018report}. The annotations for this data were collected using the Amazon Mechanical Turk platform. For each sentence, ten non-native and ten native English speakers were asked to mark words or phrases as complex. Words and phrases were annotated as complex (label \texttt{1}) if at least one of the $20$ annotators annotated them as such, and simple (label \texttt{0}) otherwise. The winning system for the binary track was a feature based ensemble approach \cite{gooding2018camb}. The system used traditional features associated with word complexity such as the frequency, word length and psycholingustic attributes of words. The performance on this dataset was subsequently improved using a neural sequence labelling approach by \citet{gooding2019complex}.

The prior datasets and systems largely focused on binary approaches to modelling CWI. Recognising the limitation of viewing word complexity in this way, \citet{shardlow2021semeval} arranged a shared task on lexical complexity prediction based on the \citet{shardlow-etal-2020-complex} dataset. The dataset contains judgements on word complexity using a Likert scale of $1$--$5$, allowing an indication of the annotators' familiarity with the term. The framing of word complexity in a continuous fashion allows for more subjective approaches of CWI to be introduced. However, overall judgements of difficulty in this dataset are still produced by averaging the labels across multiple annotators. Whilst the data gives an insight into complexity as a continuous measure, it does not provide insight into the individual differences across annotators. 

Closest to our work, \citet{gooding2021word} make the case for audience specific CWI by showing that the best models for a given proficiency group are produced using annotations from the target audience. They use proficiency annotations from the \citet{yimam-EtAl:2017:RANLP} dataset and find that the features contributing to word complexity differ depending on the audience background. This work illustrates the importance of considering the target audience in CWI. However, the populations investigated still represent coarse groups, such as native vs non-native, or beginner, intermediate and advanced. Whilst this work is a step in the right direction for subjective complexity, there still remain many differences in complexity judgements even within such groups.

Finally, a similar task to CWI is that of vocabulary prediction, whereby the size and extent of a learner's vocabulary is estimated \cite{meara1987alternative, laufer1999vocabulary}. Whilst word difficulty is subtly different to vocabulary size, as a learner may recognise a word but still find it difficult to understand, the areas are closely aligned. Empirical studies have found strong correlations between receptive vocabulary size tests and reading comprehension tests for ESL readers with learners from different proficiency levels \citep[e.g.][]{laufer1992much, qian1999assessing, qian2002investigating, henriksen2004relationship}. In fact, vocabulary size has been found to predict as much as 72\% of the variance in reading, leading many researchers to emphasise that receptive vocabulary size is the determinant factor for reading success in L2~\cite{staehr2008vocabulary}. Therefore, being able to predict a readers' vocabulary size, with a small number of samples, is very useful. The effectiveness of active learning for vocabulary prediction was investigated by \citet{ehara2014formalizing}, who showed that graph-based active learning improves vocabulary prediction for Japanese speakers. However, the active learning process is simulated and not performed in real-time with participants. 

\subsection{Active Learning}
Annotated data can be time consuming and expensive to obtain, notably in specialised domains where only experts can provide reliable labels \cite{konyushkova2017learning}. By enabling a classifier to interactively query data points, active learning allows machine learning classifiers to achieve higher accuracies with fewer training instances. It is well-motivated in many modern machine learning problems where data may be abundant, but labels are scarce or expensive to acquire \cite{settles2009active}. For instance, active learning has been used in text classification \cite{hoi2006batch}, information extraction \cite{settles2008analysis} and financial applications \cite{gooding-briscoe-2019-active}.

The task of personalised CWI is a good candidate for active learning for two reasons:

\paragraph{Labelling time} Previous CWI datasets contain thousands of annotated instances to train word complexity models. It is unfeasible to collect this amount of data, in a reasonable time frame, for an individual reader.

\paragraph{Learner lifetime}
In complex word identification we can expect the needs of a user to change over their `learner lifetime'. Whilst a model may be effective for a reader at one point in time, as their proficiency improves, the model will require retraining.

\subsubsection{Uncertainty Sampling}
In uncertainty sampling an active learner queries the instances it is least certain about how to label. This is implemented using an uncertainty measure and is one of the most commonly used active learning frameworks \cite{lewis1994sequential,settles2009active}. For many real-world learning problems, large collections of unlabelled data can be gathered easily. For instance, in our setting the unlabelled data represents all possible words. This motivates an uncertainty sampling approached called pool-based sampling \cite{lewis1994sequential} which assumes that there is a small set of labelled data ${\mathcal{L}}$ and a large pool of unlabelled instances $\mathcal{U}$. Instances are drawn from this unlabelled pool for labelling in a greedy fashion, according to an informativeness measure calculated by applying the classifier and measuring the uncertainty. 

In our paper, we employ active learning for the task of complex word identification. We train classifiers to identify whether a word would be considered difficult for a reader based on $23$ word familiarity questions. 
\section{Experiment Design}
In our experiment, participants are presented with vocabulary and asked whether they could confidently define the meaning. There are two stages to the data collection; initially, word annotations are requested in tandem with a classifier as part of an active learning process. During this stage, participants are shown a total of $23$ words to annotate.\footnote{$23$ annotations were found to be the minimum number required for model convergence during trial runs.} These annotations are used to train a personal word complexity model. Following this, annotations are collected for a test set, per participant, which can be used to validate the trained classifier. The test set contains $22$ words across CEFR levels and is presented in a randomised order. Annotators are not aware of the switch from training to test data collection. The study was hosted online and implemented using the Flask framework for Python \cite{grinberg2018flask}.

\subsection{Participants}
We collected annotations via the Prolific platform from $1{,}255$ ESL readers spanning $57$ different first languages.\footnote{\url{https://prolific.co}} Participants were required to fill out a demographic questionnaire which is available to view in our data repository. The questionnaire was informed using guidelines on judging reading ability \cite{acheson2008new} and included the self-reported English proficiency, native language, hours spent reading per week and highest level of formal education. Demographic statistics are available in Appendix~\ref{app:demographic}.

\subsection{Materials}
\begin{figure*}[t]
\centering
  \includegraphics[width=13cm]{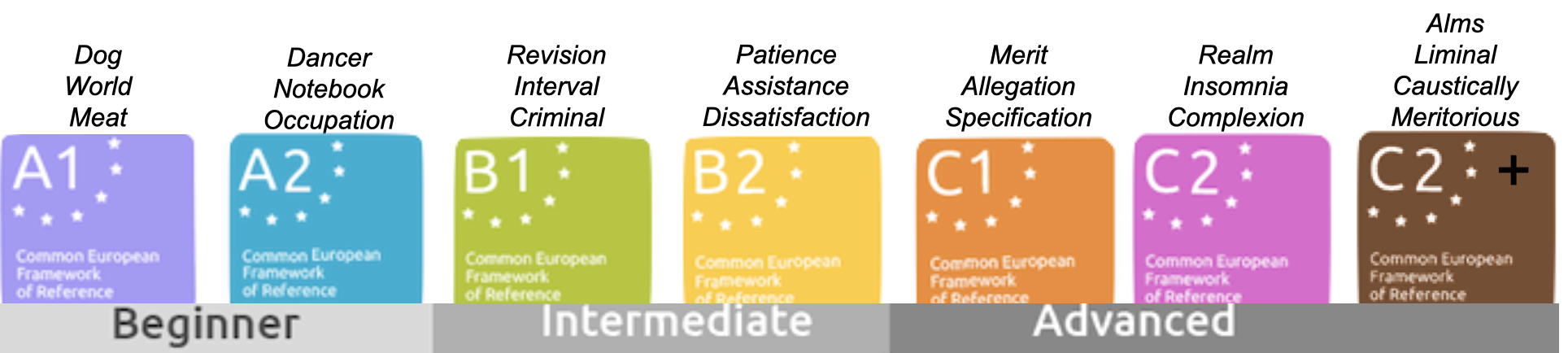}
  \caption{Test set}
  \label{fig:test_set}
\end{figure*}

We created a test set containing $22$ words spanning different levels of the Common European Framework of Reference for Languages (CEFR)~\cite{cefr} shown in Figure \ref{fig:test_set}. The items were selected by sampling those with the fewest dictionary entries to minimise word ambiguity. 

To bootstrap the active learning process, we train an initial logistic regression classifier with $150$ instances from the CWI dataset of \citet{yimam-EtAl:2017:RANLP}. The features used are those found to correlate most with non-native judgements of word complexity according to \citet{gooding2021word}, namely the word length, word frequency and psycholinguistic properties of the words such as the familiarity, concreteness, and imagability, which have been shown as important when predicting word complexity~\cite{carroll1973word,begg1969concreteness,zevin2002age}. This classifier is then tuned to the individual during the active learning process. 
\begin{figure*}[t]
 \begin{subfigure}[b]{0.33\textwidth}
                \centering
                \includegraphics[width=0.9\linewidth]{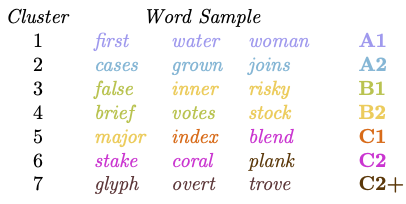}
                \vspace*{3mm}
                 \caption{Sample of words from clusters \\ \hphantom{testing testing}}
                \label{fig:cluster_examples}
        \end{subfigure}%
        \begin{subfigure}[b]{0.33\textwidth}
                \centering
                \includegraphics[width=0.9\linewidth]{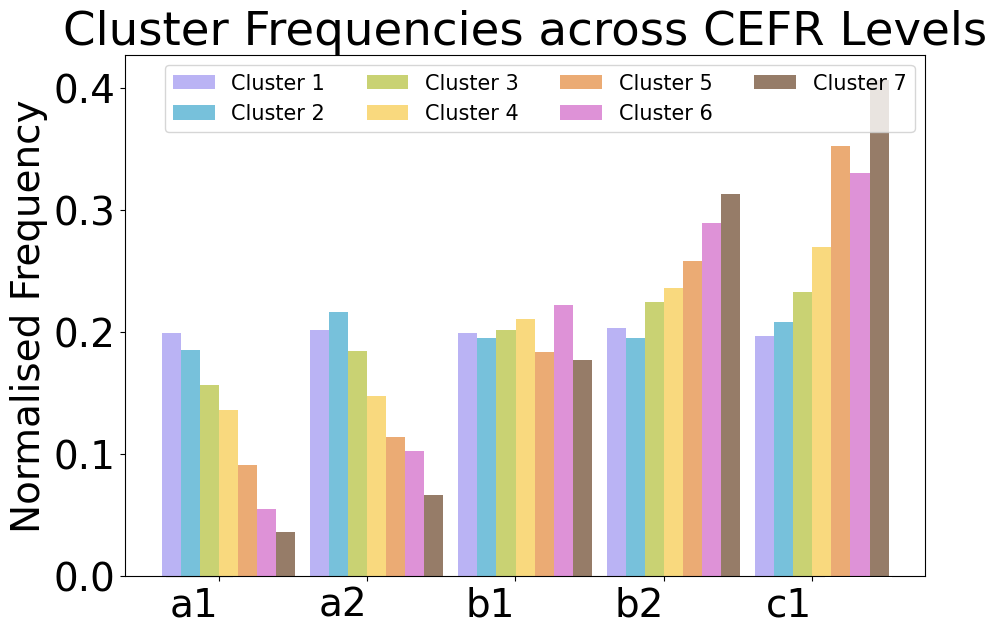}
                \caption{Frequency of vocabulary across \\ levels per cluster present in \textsc{EFLLex}}
                \label{fig:word_cluster_freq}
        \end{subfigure}%
        \begin{subfigure}[b]{0.33\textwidth}
                \centering
                \includegraphics[width=0.9\linewidth]{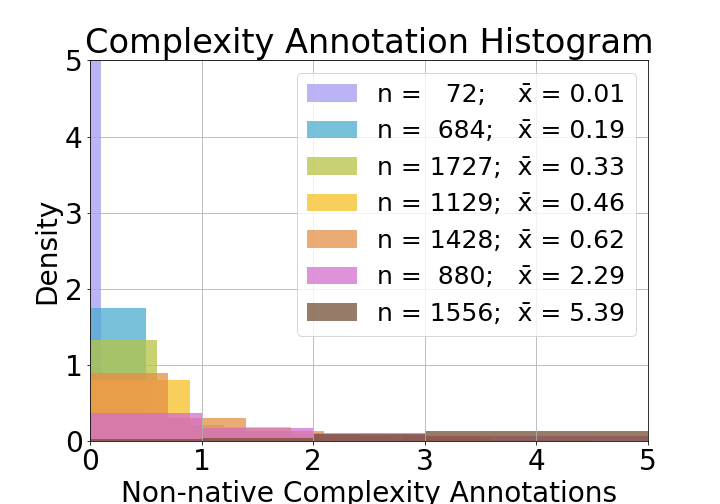}
                \caption{Histogram showing the density \\ of annotations present per cluster}
                \label{fig:histogram}
        \end{subfigure}
        \caption{Clustering statistics}
\end{figure*}
\subsubsection{Pre-clustering}
A typical approach for collecting annotations, in an active learning setting, is to select samples close to the classification boundary. However, it has been shown that taking into account the prior data distribution can improve performance \cite{settles2009active}. One way to achieve this, is by modelling the input distribution using clustering, and propagating label information to instances in the same cluster \cite{nguyen2004active}. In our experiment, we use the \citet{yimam-EtAl:2017:RANLP} dataset as our unlabelled pool and perform clustering on the $7{,}476$ uniquely labelled words prior to active learning. 

We use agglomerative clustering, a hierarchical clustering approach which performs clustering in a bottom up fashion \cite{gowda1978agglomerative}. Each word initially belongs to its own cluster and the algorithm successively merges these to form larger groups. This type of unsupervised clustering has been used across multiple natural language processing tasks including word clustering, co-reference resolution and word sense disambiguation \cite{chen2010graph}. The features used for clustering are the same as employed for the task of the complexity prediction. 

We set the number of clusters to $7$ in an attempt to align with the CEFR categories chosen in our test set. One disadvantage of hierarchical clustering methods, is that they have large storage requirements, and so can be computationally intensive to compute \cite{gan2020data}. However, as we pre-calculate and cache clusters the time complexity is that of base informativeness measures, e.g.\ uncertainty sampling \cite{settles2008analysis}. The time complexity is a necessary consideration, as we are conducting active learning interactively with oracles in real-time. 

An example of words from the resulting clusters is displayed in Figure \ref{fig:cluster_examples}. Words of length $5$ were sampled randomly and have been colour coded according to the corresponding CEFR level.\footnote{CEFR levels were obtained via \url{https://languageresearch.cambridge.org/wordlists/text-inspector}}
\subsubsection{Cluster Evaluation}
We evaluate our clusters using two techniques; the first employs \textsc{EFLLex} which is a graded lexical resource for learners of English as a foreign language. The dataset, collected by \citet{durlich2018efllex}, contains $15{,}280$ entries from $1{,}971$ graded texts. The frequency of the lemma at each level of the CEFR (C2 excepted) is provided. A total of $5{,}983$ entries were present in both \textsc{EFLLex} and our clusters. We plot the average frequencies of words in each cluster across CEFR levels in Figure \ref{fig:word_cluster_freq}. Generally, the plot shows that there are clear differences in the levels of vocabulary across clusters. When we consider A1 texts, the cluster with the highest average frequency of words is Cluster $1$. There is a linear decrease in the frequency of words at this level as the cluster level increases. When we consider the proportion of higher level vocabulary (i.e.\ B1 to C2), a clear trend emerges where Clusters $5$--$7$ contain words that appear much more frequently at this level. Cluster $7$, contains words at the highest level of CEFR for instance \textit{glyph}, and is shown to contain words that occur much more commonly at C1 and B2 level. 

The second way we evaluate clusters is using the annotations of the \citet{yimam-EtAl:2017:RANLP} dataset. As we use this data for our unlabelled pool, we have the complexity annotations for each item from both native and non-native annotators. Our work focuses on complexity prediction for ESL readers and so we evaluate clusters using the non-native annotations. This results in a complexity score per word ranging from $1$--$10$ which represents how many non-native annotators considered the word to be complex. A histogram showing the density of non-native complexity annotations present for words in each cluster is shown in Figure \ref{fig:histogram}. The average of these scores per cluster ($\bar{x}$) is shown in Figure \ref{fig:histogram}, as well as the number of words present in each cluster ($n$). Clusters are colour coded according to the same key in Figure \ref{fig:histogram}. Cluster $7$ contains the highest average complexity labels ($5.39$) and Cluster $1$ the lowest ($0.01$). As the cluster levels increase, we see a shift in the density of complexity annotations present, supporting the notion that these clusters contain vocabulary at differing levels. 

\subsection{Active Learning Loop}
\begin{figure}[t]
\centering
  \includegraphics[width=6cm]{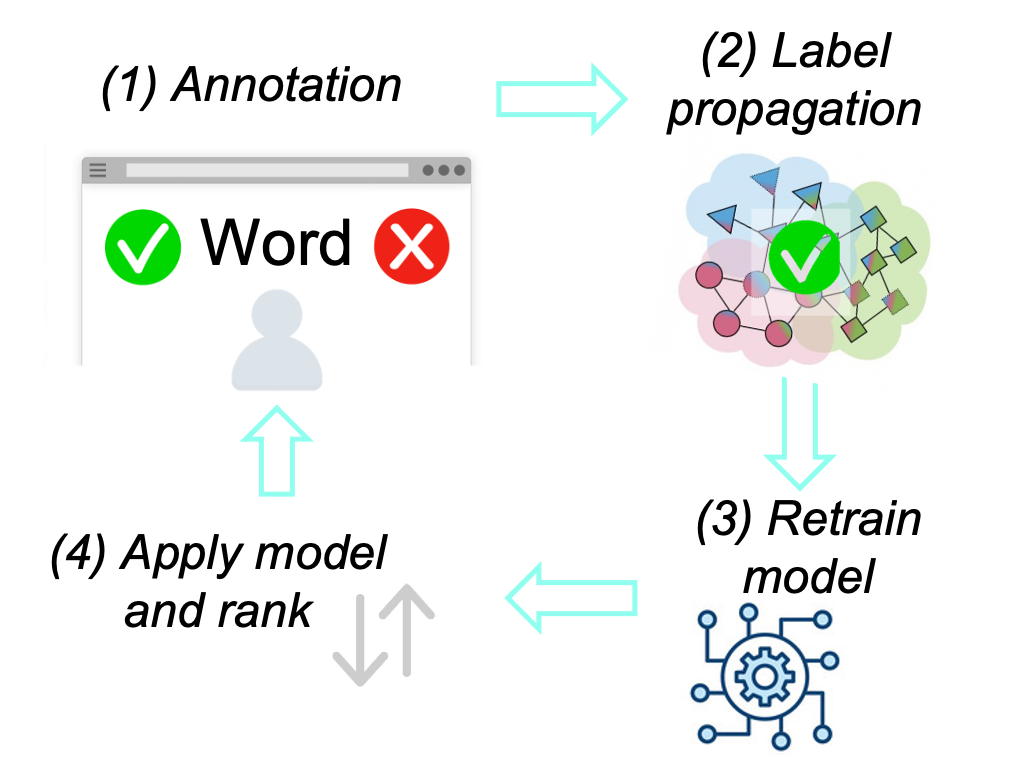}
  \caption{Figure showing the stages of our active learning framework}
  \label{fig:experiment_loop}
\end{figure}
Figure \ref{fig:experiment_loop} shows the four stages that take place during the active learning process. Each of these steps is described below. 

\paragraph{(1) Annotation}
 In the initial step, the annotator is presented with a word form and asked whether they could confidently define the meaning. Whilst the context of a word impacts its meaning and complexity, we use word forms as this allows for a pool-based active learning approach. Furthermore, the use of word forms alone has been shown as an effective measure of proficiency and vocabulary prediction \cite{lemhofer2012introducing, ehara2014formalizing}. The user interface showing the instructions for this experiment are provided in Appendix~\ref{app:interface}. When the annotator selects an answer this provides a binary annotation for the word. 
 
\paragraph{(2) Label propagation}
The chosen label is then propagated to the closest $150$ instances from the clustered pool. 

\paragraph{(3) Re-train model}
We train a log-linear model (i.e.\ logistic regression) per individual, for the following reasons: the model can provide a class ownership probability which is required for active learning, log-linear models are flexible and interpretable, finally the algorithm is efficient and able to train in $\mathcal{O}(nd)$ time, where $n$ is the number of samples and $d$ the feature dimensions \cite{gujarati2009measure,bulso2019complexity}. As each model is re-trained every time an annotation is provided, the speed of training was a critical consideration. %An initial seed classifier is trained using $150$ binary annotations from the \cite{yimam-EtAl:2017:RANLP} dataset.  

\paragraph{(4) Apply model and re-rank}
Once the model has been re-trained, it is applied to the unlabelled pool so that the samples can be ranked according to the model uncertainty. The uncertainty sampling technique we use is entropy \cite{shannon1948mathematical}. In the following notation $x^{*}$ denotes the most informative instance from an unlabelled vocabulary pool:
\begin{equation*}
    x^{*}_{H} = \argmax_{x} - \sum_{i} P_{\theta}(y_{i}|x)\log P_{\theta}(y_{i}|x)
\end{equation*}
where $y_{i}$ ranges over the possible labellings. The instance from the pool that the model is least certain about is then presented to the annotator for the next annotation step.

\section{Results}
In our experiment, each participant produces a gold test set containing annotations for words across CEFR levels. This varies for each individual, as word complexity is subjective. For instance, consider the test set samples shown in Table~\ref{tab:test_examples}; words marked with a $1$ have been annotated as difficult by participants. Participant $a$, who is self-reported to be intermediate, has found three words in this sample difficult, whereas participant $b$, self-reported at a near native level, only reports one. 

\begin{table}[t]
\begin{tabular}{lcccc} \toprule
   & \multicolumn{1}{l}{{\textit{World}}} & \multicolumn{1}{l}{{\textit{Realm}}} & \multicolumn{1}{l}{{\textit{Merit}}} & \multicolumn{1}{l}{{\textit{Alms}}} \\ \hline
{$P_a$} & {$0$}                                  & {$1$}                                  & {$1$}                                  & {$1$}                                 \\
{$P_b$} & {$0$}                                  & {$0$}                                 & {$0$}                                 & {$1$} \\\bottomrule                               
\end{tabular}
\caption{Test set samples from participants $P_a$ and $P_b$}
\label{tab:test_examples}
\end{table}

We calculate the F-score and Kappa agreement across all annotations to evaluate the performance of bespoke models. The F-score represents the harmonic mean between the precision and recall of the classifier, whereas the Kappa coefficient measures pairwise agreement, correcting for expected chance agreement \cite{cohen1960coefficient}. 

\begin{table*}[t]
\begin{tabular}{lcccccc} \toprule
\textbf{Model} & \multicolumn{6}{c}{\textbf{Proficiency level}}                                                                                                                                          \\
                & \multicolumn{1}{c}{All} & \multicolumn{1}{c}{Beginner} & \multicolumn{1}{c}{Intermediate} & \multicolumn{1}{c}{Advanced} & \multicolumn{1}{c}{Near Native} & \multicolumn{1}{c}{Native} \\ \midrule
                & \multicolumn{6}{c}{\textit{F-score} }                                                                                                                                                         \\   
Active Learning & \textbf{0.751}          & \textbf{0.763}               & \textbf{0.720}                   & \textbf{0.752}               & \textbf{0.764}                  & \textbf{0.761}             \\
Group Average   & 0.669                   & 0.608                        & 0.624                            & 0.679                        & 0.726                           & 0.754                      \\
Seq-CWI         & 0.677                   & 0.620                        & 0.688                            & 0.689                        & 0.660                           & 0.642                      \\
Frequency       & 0.534                   & 0.461                        & 0.529                            & 0.541                        & 0.530                           & 0.514                      \\
All Simple      & 0.441                   & 0.450                        & 0.436                            & 0.437                        & 0.447                           & 0.454                      \\ \midrule
                & \multicolumn{6}{c}{\textit{Kappa}}                                                                                                                                                             \\ 
Active Learning & \textbf{0.503}          & \textbf{0.525}               & \textbf{0.441}                   & \textbf{0.504}               & \textbf{0.529}                  & \textbf{0.524}             \\
Group Average   & 0.379                   & 0.296                        & 0.304                            & 0.395                        & 0.473                           & 0.523                      \\
Seq-CWI        & 0.395                   & 0.300                        & 0.409                            & 0.414                        & 0.371                           & 0.347                      \\
Frequency       & 0.189                   & 0.083                        & 0.173                            & 0.194                        & 0.196                           & 0.182                       \\ \midrule   \textit{Test size}       & 26,136                   & 176                        & 4,180                            & 12,122                        & 7480                           & 2,156           \\ \bottomrule
\end{tabular}
\caption{Model results grouped by annotator proficiency}
\label{tab:results_sota}
\end{table*}

\begin{table}[t]
\begin{tabular}{lcc} \toprule
                                                 \textbf{Model}           &  \multicolumn{1}{l}{\textit{F-score}} & \multicolumn{1}{l}{\textit{Kappa}} \\ \midrule
\begin{tabular}[c]{@{}l@{}}Random \end{tabular}  & 0.605                                & 0.218                              \\
\begin{tabular}[c]{@{}l@{}}Cluster\end{tabular} & 0.649                                & 0.361                              \\
\begin{tabular}[c]{@{}l@{}}AL\end{tabular}  & 0.740                                & 0.482  \\ \bottomrule                           
\end{tabular} 
\caption{Results showing differing sampling techniques applied to $2{,}200$ annotations from $100$ participants}
\label{tab:sampling}
\end{table}
We additionally run an experiment to evaluate the effectiveness of active learning compared to randomly sampling vocabulary. In the experiment, we require annotators to label a sample of $22$ random words, as well as $22$ words sampled using active learning. As this doubles the annotation effort we perform this with a subset of $100$ annotators. The results are shown in Table \ref{tab:sampling}. We compare with models trained using random sampling as well as random sampling with cluster-based label propagation. The best results are achieved using the active learning framework. 

We evaluate the performance of the active learning models and compare to the following baselines. (1) Group Average: we calculate the average of the test set complexity annotations across participants, if a word has been labelled as complex by more than 10\% of the target group it is labelled as difficult. This is a competitive baseline as we use the direct annotations of the target audience on the test sets. (2) Seq-CWI: we compare to a state-of-the-art system for complex word identification \citep{gooding2019complex}. However, we note that this is an unfair comparison as the system is one-size-fits-all and not audience specific. (3) Frequency: relies on frequency thresholding, where all words below a given frequency are marked as complex. (4) All Simple: all words are marked as simple.

Table~\ref{tab:results_sota} shows the results of the active learning models compared to baselines. We group results according to the proficiency of participants. The beginner results are shown for completeness, but should be interpreted with caution due to the smaller test set size ($176$). Kappa values in the range of $0.40$--$0.60$ represent moderate agreement \cite{cyr1992measures}, although it has been argued that scores above $0.45$ represent substantial agreement \cite{munoz1997interpretation}.  

When we consider the active learning models, we see that the F-score calculated across all test annotations is $0.751$, and the Kappa agreement is $0.503$. The active learning models outperform the competitive group-average baseline which supports the hypothesis that individual models are preferable to aggregate group-level approaches. The best scores for all proficiency groups are achieved using the active learning framework. However, for the native annotations, the difference between the group-based approach and active learning models is small. When considering the group-average baseline we see that both the F-score and Kappa values increase in line with proficiency. This supports the notion that the concept of word complexity stabilises for higher proficiency levels \cite{krstic2018all,gooding2021word}. As such, aggregate models may work well for highly proficiency audiences and individual models best for lower levels where word knowledge is more variable. 

% Both of these results show an increase on the baselines as well as the current sota system. The best results are obtained when predicting for the \textit{near native} audience, following this the \textit{native} and \textit{advanced}. Figure \ref{fig:fscoredist} shows the dispersion of F-scores across participant groups including both the frequency baseline and sota systems.

% \begin{figure}[!t]
% \centering
%   \includegraphics[width=\textwidth]{images/fscores_dist.png}
%   \caption{Plot showing dispersion of F-scores across proficiency subgroups, as well as all model scores from the active learning, sota and frequency threshold systems.}
%   \label{fig:fscoredist}
% \end{figure}

\subsection{Analysis}
\begin{table*}[t]
\centering
\begin{tabular}{llllllllll}
\multicolumn{5}{r}{\small{Near Native:}}                         & \multicolumn{1}{c}{\small{0.34}} &      & \multicolumn{1}{c}{\small{0.32}} &      &                                 \\
\multicolumn{5}{r}{\small{Advanced:} }                           & \multicolumn{1}{c}{\small{0.51}} &      & \multicolumn{1}{c}{\small{0.47}} &      &                                 \\
\multicolumn{5}{r}{\small{Intermediate:}  }                      & \multicolumn{1}{c}{\small{0.59}} &      & \multicolumn{1}{c}{\small{0.59}} &      &                                 \\ \hline
\multicolumn{1}{|l}{(1)  The} & current & cuts & in & public & expenditure              & will & exacerbate               & this & \multicolumn{1}{l|}{situation.} \\
\multicolumn{10}{|c|}{$\downarrow$}                                                                                                                              \\
\multicolumn{1}{|l}{(2)  The} & current & cuts & in & public & \textit{spending}        & will & exacerbate               & this & \multicolumn{1}{l|}{situation.} \\
\multicolumn{1}{|l}{(3)  The} & current & cuts & in & public & \textit{spending}        & will & \textit{worsen}          & this & \multicolumn{1}{l|}{situation.} \\ \hline
\end{tabular}
\caption{Example of how simplification output changes depending on audience specific CWI models. The complexity probability for audiences is the result of aggregating output from all trained models across proficiencies. A word is considered complex if the probability is above $0.50$.}
\label{tab:simplification_example}
\end{table*}
\begin{table}[t]
\centering
\begin{tabular}{llllll}
                                                                                      & \multicolumn{5}{c}{\small{\textit{}}} \\ \toprule
\multicolumn{1}{l}{}                                                                 & \small{A1}     & \small{A2}     & \small{B1}      & \small{B2}      & \small{C1}      \\ \midrule
\multicolumn{1}{l}{\small{{Intermediate}}} & \small{$333$}    & \small{$881$}    & \small{$1208$}    & \small{$2109$}    & \small{$2474$}    \\
\multicolumn{1}{l}{\small{{{Advanced}}}}                                                         & \small{$277$}    & \small{$754$}    & \small{$1016$}    & \small{$1773$}    & \small{$2062$}    \\
\multicolumn{1}{l}{\small{{Near Native}}   }                                                   & \small{$232$}    & \small{$643$}    & \small{$866$}     & \small{$1505$}    & \small{$1735$}   \\ \bottomrule
\end{tabular}
\caption{The average number of words identified as complex across CEFR levels. Results are averaged from a sample of $100$ classifiers from each proficiency group.}
\label{tab:cefr_complex_preds}
\end{table}

Each annotator produces a bespoke model based on their personal perception of word difficulty. We tested whether there was a significant difference in complexity predictions of models depending on the annotator's self-reported proficiency. To investigate this, we selected $100$ models from near native, advanced and intermediate annotators. We control for the number of classifiers due to having an unequal number of annotators across levels. 

These models are applied to the \textsc{EFLLex} dataset, removing any vocabulary that may have been seen during train time. This results in complexity predictions for approximately $10{,}000$ words per model. The average number of words found complex by these classifiers is shown in Table \ref{tab:cefr_complex_preds}. The intermediate classifiers, on average, identify more complex words across all vocabulary levels, whereas the near native classifiers predict fewer words as difficult across all levels compared to both the intermediate and advanced classifiers. We calculate the significance using \citeauthor{Satterthwaite}’s method \cite{kuznetsova2017lmertest}, applied to a mixed-effects model that treats the first language of annotators and the trained model as random effects.\footnote{Using R formula notation, the model is: $\mathrm{model\_predictions} \sim \mathrm{annotator\_proficiency}
+ {(1\mid\mathrm{annotator\_L1})} + {(1\mid\mathrm{model})}$} 
We find that the difference between model predictions depending on the proficiency of the annotator is highly significant ($p <1{\times}10^{-10}$). 

\subsection{Downstream Tasks}
Being able to distil the vocabulary experience of an individual, using a limited number of annotations, is extremely useful for many applications. For instance, when judging the proficiency of a reader. Using our trained classifiers, we run a preliminary experiment to see if it is possible to predict the proficiency of the annotator using their model predictions. We train a linear model to predict whether the annotator self-reported at an intermediate, advanced or near native level. We use the total number of C1 words identified as complex as a feature and test using 5-fold stratified cross-validation. By training a simple model with this feature alone, we can obtain a precision of $0.70$. We release the trained models with our paper to encourage further work on the applications and evaluation of personal CWI models for proficiency prediction. 

Finally, a further application that benefits from individualised models of word complexity is text simplification. To illustrate, we provide an example in Table \ref{tab:simplification_example} which was produced using an open source lexical simplification system by \citet{gooding2019recursive}. The system is designed to accommodate differing models of word complexity so that simplification output can be personalised. In this example, we use the mean probability that a word is complex by averaging the output from our classifier groups. Sentence (1) shows the result of using near native CWI models, no words are considered complex and so no simplifications are performed. The advanced models produce a probability larger than $0.50$ for the word \textit{expenditure} and it is therefore simplified. Finally, the intermediate models result in two words being identified as complex which are subsequently simplified. This example provides further motivation for why considering word complexity in a subjective manner is beneficial. 

\section{Conclusions and Future Work}
We show that with active learning it is feasible to build individual models of word complexity. Our models are able to be trained efficiently, in real-time, using only $23$ word annotations per individual. Models trained in this fashion are able to accommodate the subjective nature of word complexity, thereby facilitating personalised readability and simplification systems.
We release our dataset containing $55{,}125$ word complexity annotations, participant metadata and models, providing a new benchmark for the task of human-centered word complexity prediction. 

Our work is a first step towards the development of personal complexity models. In future work, we aim to more extensively investigate the relationship between a learner's perceived and actual knowledge of words, as it has been reported that there is a misalignment between what a learner thinks they know, and what they are actually able to define \cite{laufer1997lexical}. Additionally, the context of a word impacts its perceived difficulty \cite{gooding:2019, strohmaier-etal-2020-secoda}; considering token-level complexity as opposed to type-level complexity is another important avenue for future work. \newpage

\bibliography{anthology,custom}
\bibliographystyle{acl_natbib}
\newpage
\appendix

%\section*{Appendices}

\section{Participant demographics}
\label{app:demographic}
The demographic information of participants is shown in Table \ref{tab:participants}. A more detailed breakdown including first language statistics is available on our data repository. 
\begin{table*}[t]
\begin{tabular}{ll|ll|ll|ll}
\multicolumn{2}{c}{Proficiency (\%)} & \multicolumn{2}{c|}{Education (\%)} & \multicolumn{2}{c|}{Age range (\%)} & \multicolumn{2}{c}{\begin{tabular}[c]{@{}c@{}}Hours reading English\\ per week (\%)\end{tabular}} \\ \hline
Native               & 8.29         & Graduate                & 31.16     & 18 - 24           & 39.24           & 0 - 10                                            & 40.40                                          \\
Near native          & 30.50         & Undergraduate           & 29.83     & 25 - 34           & 22.17          & 10 - 20                                           & 22.83                                          \\
Advanced             & 45.51        & High School             & 33.37      & 35 - 44           & 14.52           & 20 - 30                                          & 14.95                                          \\
Intermediate         & 15.14         & Vocational Training     & 5.42      & 45 - 54           & 11.64           & 30 - 40                                          & \hphantom{d}9.82                                           \\
Beginner             & 0.66          & No formal education     & 0.22      & 55+               & \hphantom{d}9.53           & 40 +                                             & 11.98                                         
\end{tabular}
\caption{Background statistics for the $1225$ participants}
\label{tab:participants}
\end{table*}

\section{Experiment interface}
\label{app:interface}
\begin{figure*}[t]
\centering
  \includegraphics[width=7cm]{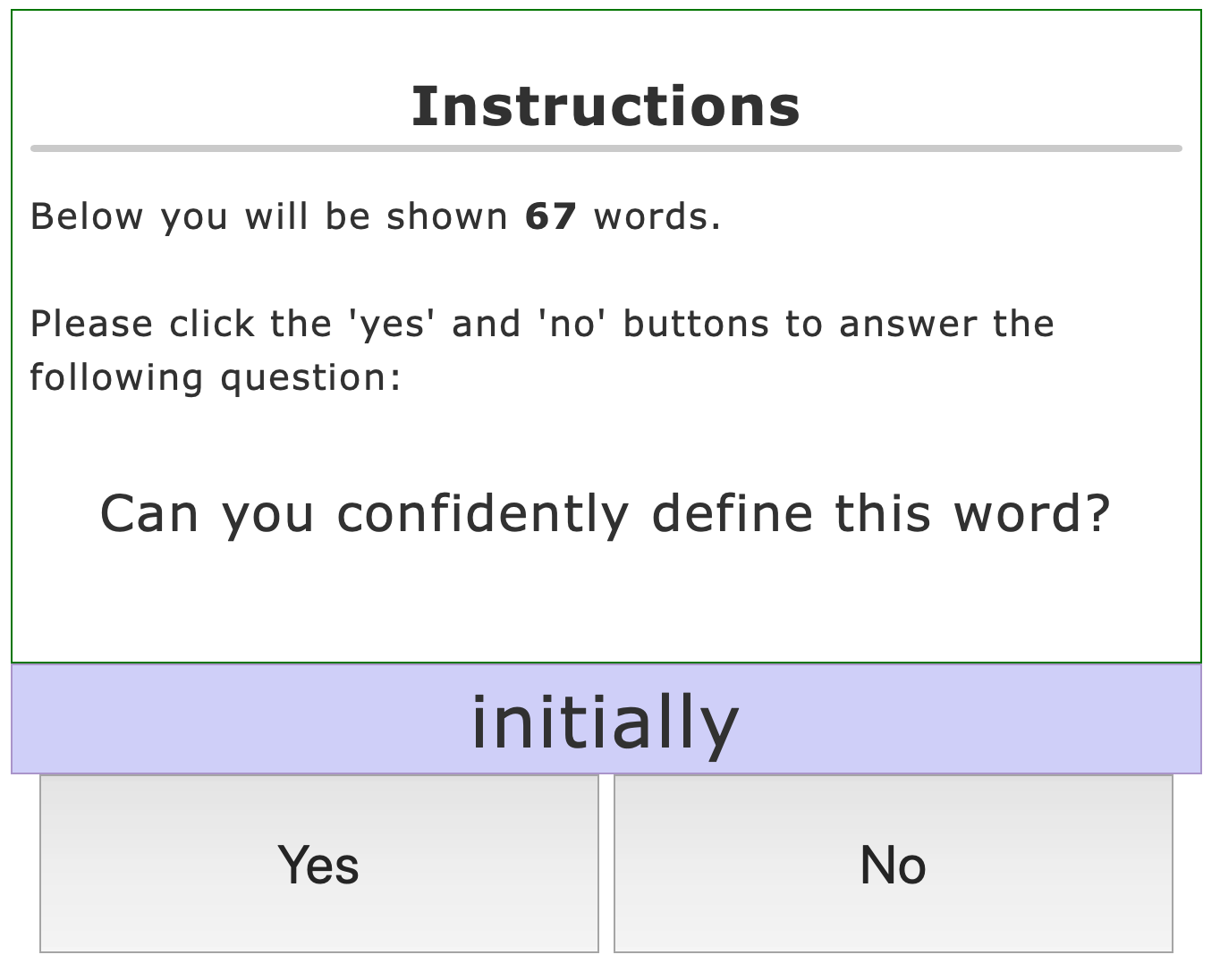}
  \caption{Experiment interface}
  \label{fig:experiment}
\end{figure*}

The experiment interface is shown in Figure \ref{fig:experiment}. Annotators were paid 15 cents per minute in accordance with Prolific guidelines. Annotators were informed that the study and their subsequent annotations were for research purposes. Data is fully anonymised and both this and the models will be released under the Affero General Public License 3 (AGPL-3.0). All participant instructions as well as the demographic questionnaire are included in our GitHub data repository.

\end{document}